%% file: main.tex
\documentclass[10pt,twocolumn,letterpaper]{article}

\usepackage{iccv}              


\usepackage{multirow}
\usepackage{color}
\usepackage{xcolor}
\usepackage{colortbl}
\usepackage{booktabs}
\usepackage{stackengine}
\usepackage{bbm}
\usepackage[table, dvipsnames]{xcolor}
\usepackage{xcolor}

\definecolor{wrong}{rgb}{.8,.349,.1}
\definecolor{right}{rgb}{.3,.7,.1}
\definecolor{Gray}{gray}{0.9}	
\definecolor{mygrey}{rgb}{0.949,0.949,0.949}
\newcolumntype{a}{>{\columncolor{mygrey}}c}

\definecolor{iccvblue}{rgb}{0.21,0.49,0.74}
\usepackage[pagebackref,breaklinks,colorlinks,allcolors=iccvblue]{hyperref}

\def\paperID{2112} 
\def\confName{ICCV}
\def\confYear{2025}

\title{MedVSR: Medical Video Super-Resolution with Cross State-Space Propagation}

\author{Xinyu Liu$^{1}$, Guolei Sun$^{2}$, Cheng Wang$^{1}$, Yixuan Yuan$^{1,}$\thanks{Corresponding authors.~Prof.~Yuan was supported by Hong Kong Innovation and Technology Commission Innovation and Technology Fund ITS/229/22, CUHK 4055269, Hainan Province Clinical Medical Center.} ~, Ender Konukoglu$^{2,}$\footnotemark[1]~\\ 
$^1$ The Chinese University of Hong Kong,  $^2$ Computer Vision Laboratory, ETH Zurich\\
}

\begin{document}
\maketitle
\input{sec/0_abstract}    
\input{sec/1_intro}
\input{sec/2_related}
\input{sec/3_finalcopy}
\input{sec/4_experiments}
\input{sec/5_conclusion}
{
    \small
    \bibliographystyle{ieeenat_fullname}
    \bibliography{main}
}

\end{document}

%% file: sec/0_abstract.tex
\begin{abstract}
High-resolution (HR) medical videos are vital for accurate diagnosis, yet are hard to acquire due to hardware limitations and physiological constraints. 
Clinically, the collected low-resolution (LR) medical videos present unique challenges for video super-resolution (VSR) models, including camera shake, noise, and abrupt frame transitions, which result in significant optical flow errors and alignment difficulties. Additionally, tissues and organs exhibit continuous and nuanced structures, but current VSR models are prone to introducing artifacts and distorted features that can mislead doctors.
To this end, we propose MedVSR, a tailored framework for medical VSR. 
It first employs Cross State-Space Propagation (CSSP) to address the imprecise alignment by projecting distant frames as control matrices within state-space models, enabling the selective propagation of consistent and informative features to neighboring frames for effective alignment.
Moreover, we design an Inner State-Space Reconstruction (ISSR) module that enhances tissue structures and reduces artifacts with joint long-range spatial feature learning and large-kernel short-range information aggregation.
Experiments across four datasets in diverse medical scenarios, including endoscopy and cataract surgeries, show that MedVSR significantly outperforms existing VSR models in reconstruction performance and efficiency. Code released at \href{https://github.com/CUHK-AIM-Group/MedVSR}{https://github.com/CUHK-AIM-Group/MedVSR}.
\vspace{-8pt}
\end{abstract}

%% file: sec/1_intro.tex
\section{Introduction}
\label{sec:intro}

\begin{figure}
    \centering
    \includegraphics[width=0.90\linewidth]{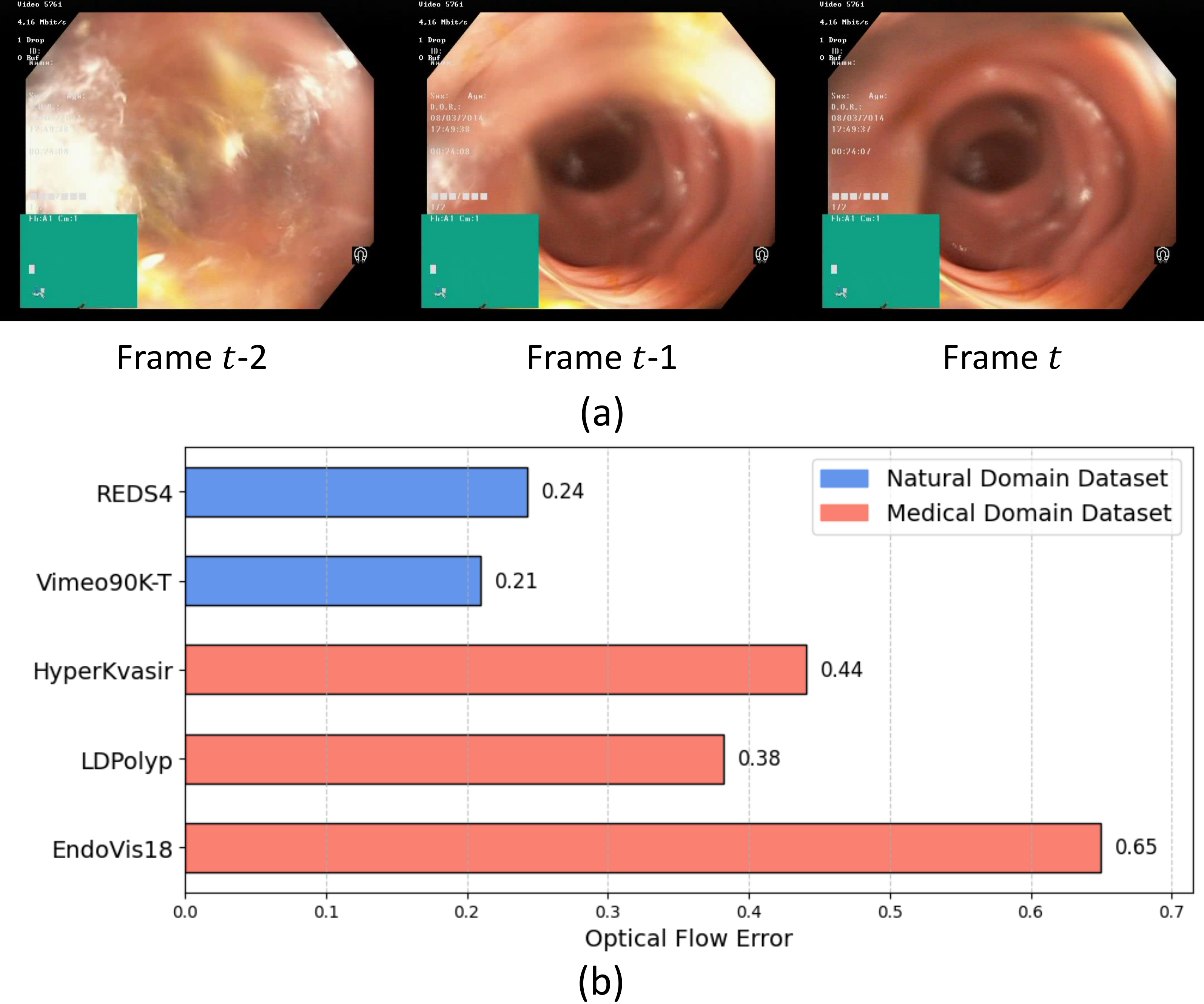}
    \vspace{-8pt}
    \caption{(a) Example of sharp transitions and jitter in medical videos, especially for distant frames (e.g. frame $t\scalebox{0.75}[1.0]{ - }2\scalebox{0.9}[1.0]{\( \rightarrow \)} t$), which pose significant challenges for existing alignment methods. (b) We measure the averaged error of forward and inverse backward optical flows in different datasets, lower error values denote better estimation stability. Medical videos tend to have significantly larger errors, making the alignment more challenging.}
    \label{fig:framejitter}
  \vspace{-12pt}
\end{figure}
In modern medical imaging, high-resolution (HR) medical videos play a critical role in providing detailed texture and structural information, which are essential for accurate diagnosis and quantitative analysis \cite{qiu2023rethinkingsrmedical}. However, obtaining HR medical videos in clinical practice is a significant challenge. The quality of medical videos is often limited by various factors, including hardware conditions, acquisition time, physiological movement, and patient comfort \cite{peters2001image, chow2016review, liu2025endogen}. As a result, medical videos frequently suffer from low-resolution (LR) and noise, which can compromise the reliability of diagnosis and treatment. 

\begin{figure}
    \centering
    \includegraphics[width=0.95\linewidth]{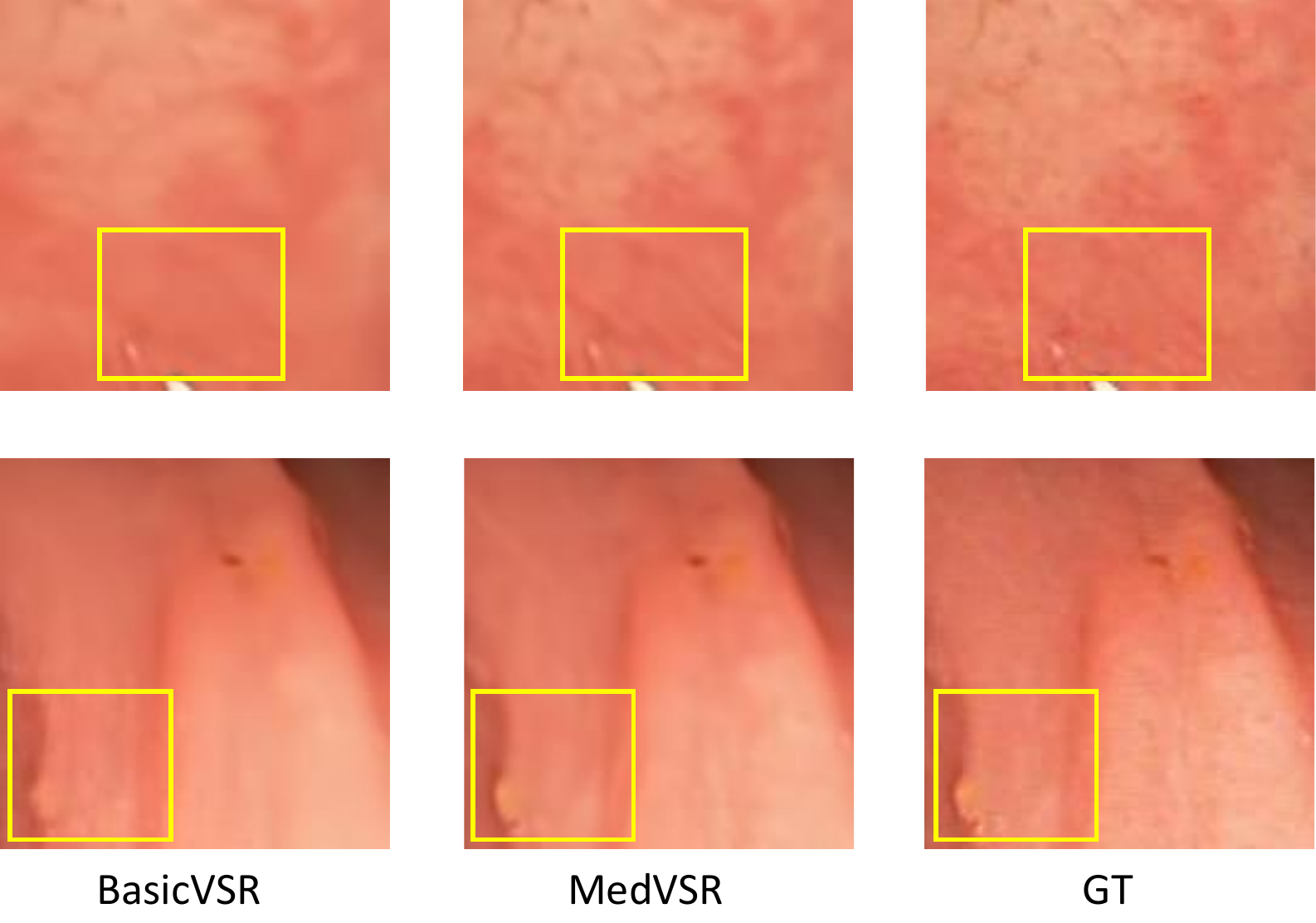}
    \vspace{-8pt}
    \caption{Examples of texture removal and shape distortion of existing VSR method \cite{chan2021basicvsr}, which could not accurately reflect the GT and can be misleading for doctors. MedVSR reconstructs real features and produces consistent results as GT. \textit{Zoom in for details}.}
    \label{fig:frameartifacts}
  \vspace{-9pt}
\end{figure}

In recent years, video super-resolution (VSR) techniques have been studied \cite{xue2019video, wang2019edvr, chan2021basicvsr, chan2022basicvsr++, xu2024iart, cao2021vsrt, liu2022tcnet, 
zhou2022video, song2022deformable, hayat2024sevsr, hafner2013pocs, almalioglu2020endol2h} to recover the HR videos from LR sequences in the natural scenes. However, in contrast to natural domain videos, medical videos pose \textit{two unique challenges}. \textbf{(1)} Medical videos captured during surgical or colonoscopy procedures often exhibit instabilities due to hardware device limitations or in-body environment constraints, such as camera shake, jitter, and sharp transitions between frames, which are illustrated in Fig.~\ref{fig:framejitter}(a). These instabilities can result in optical flow estimation errors and alignment difficulties. Compared to natural domain datasets (REDS4 \cite{nah2019ntire}, Video90K-T \cite{xue2019video}), medical videos display significantly larger optical flow errors, as shown in Fig.~\ref{fig:framejitter}(b).
This difficulty makes it crucial to develop a reliable propagation scheme between frames for more robust alignment. \textbf{(2)} Medical videos often contain tissues with continuous and nuanced structures with homogeneous intensity profiles \cite{krishnan1998intestinal}, making them highly sensitive to artifacts and reconstruction errors. Existing VSR models typically rely on stacked local CNNs with limited receptive fields \cite{chan2021basicvsr, chan2022basicvsr++, liang2022rvrt}, 
may not adequately address the reconstruction of propagated features. This limitation can lead to the introduction of artifacts, distorted object shapes, and texture removal, as shown in Fig.~\ref{fig:frameartifacts}, which compromises the reliability of the algorithm severely. The propagation of these artifacts can obscure critical details that clinicians rely on for making informed decisions.

To address the aforementioned challenges, we first propose Cross State-Space Propagation (CSSP) based on state-space models (SSM) \cite{nguyen2022s4nd, gu2023mamba, dao2024mamba2} that could effectively leverage distant frames to enhance the feature propagation process. 
Instead of directly aligning these distant frames, 
CSSP extracts and propagates reliable features by leveraging the interaction between distant and neighboring frames.
Specifically, it transforms the frame features into state-space sequences, and forwards them into a Cross State-Space Block (CSSB).
The CSSB learns a control matrix from the state-space of the distant frame and a hidden state feature from the state-space of the neighboring frame. With the recurrent selective scanning paradigm in SSM, a compound state-space is formed to produce the consistent and stable features across frames for alignment, thereby mitigating the impact of imprecise alignments of distant frames caused by sharp transitions.

Moreover, we introduce Inner State-Space Reconstruction (ISSR) that leverages the sequential processing of propagated features to enhance the inherent continuity of tissue structures in medical videos.
We concatenate the features of the current frame from the forward and backward branches, and apply SSM to extract the continuous long-range spatial feature in a recurrent manner in the Inner State-Space Block (ISSB).
Additionally, we incorporate large kernel separable blocks (LKSB) for joint short-range information aggregation during reconstruction, resulting in reduced artifacts and precise reconstructed features. 
This capability of preserving the continuous and detailed structures could potentially contribute to improved diagnostic accuracy and reliability in medical imaging applications. 

Based on the above insights, we propose MedVSR, a tailored medical video super-resolution framework that effectively addresses the challenges in improving the resolution of medical videos. We extensively validated MedVSR on multiple medical scenes, including endoscopy, surgical instruments, and cataract surgery. MedVSR demonstrates better performance and efficiency compared to state-of-the-art models. To summarize, our contributions are as follows:
\begin{itemize}
    \item We design CSSP that projects distant frames as control matrices to support the selective scanning of neighboring frames in SSM, which captures consistent features for propagation and addresses the imprecise alignment issue.
    \item We introduce ISSR that jointly learns long-range spatial features with SSM and aggregates short-range information with large-kernel convolutions, which enhances the details of reconstructed frames and reduces artifacts.
    \item We propose a MedVSR framework tailored for medical video super-resolution with CSSP and ISSR, and conduct comprehensive experiments across multiple distinct medical scenarios. MedVSR achieves state-of-the-art performance with remarkable efficiency, demonstrating its efficacy and potential for clinical application.
\end{itemize}

%% file: sec/2_related.tex
\vspace{-2pt}
\section{Related Work}
\label{sec:rel}

\vspace{-2pt}
\subsection{Video Super-Resolution}

VSR aims to restore a sequence of HR video frames from its degraded LR counterparts \cite{wang2019edvr, xue2019video, chan2021basicvsr, chan2022basicvsr++, shi2022rethinking, liang2022rvrt, xu2024iart, liu2022tcnet, yi2019progressive, lucas2019generative, yan2019frame}. Its difference with image super-resolution primarily lies in the utilization of propagated temporal information for alignment, then uses the aligned feature for reconstruction. 
Early works 
\cite{xue2019video, kim2018spatio, sajjadi2018frame} proposed to apply optical flow to estimate the motion, then propagated the warped neighboring frames for alignment. BasicVSR \cite{chan2021basicvsr} adopted alignment at feature level, and propagated the features in a bidirectional recurrent way. BasicVSR++ \cite{chan2022basicvsr++} further improved the scheme to second-order propagation, which also aligns the distant frames.
However, inaccurate flows could lead to performance degradation, especially for long-distance frames. 
To compensate for the deficiency of optical flow, recent works have proposed to learn the alignment via deformable convolution networks (DCN) \cite{chan2022basicvsr++} or deformable attention \cite{liang2022rvrt}. Meanwhile, some works \cite{shi2022rethinking, xu2024iart} have discovered that alignment in patchwise could enhance robustness. However, they all assume that aligning distant frames via higher-order propagation could effectively transmit the information, which does not strictly hold in the medical scene. In this paper, rather than propagating imprecisely aligned distant frames, we leverage them to support the learning of consistent and stable features during propagation.

\subsection{State-Space Models}

State-Space Models (SSMs) have gained considerable attention due to their efficacy in long-range sequence modeling, while requiring significantly less computational cost than full attention. Numerous models have been proposed based on SSMs. The most popular one is Mamba \cite{gu2023mamba}, which implemented a selection mechanism that makes parameters dependent on the input, all while maintaining linear complexity. Its improved version Mamba2 \cite{dao2024mamba2} proposed to produce the SSM parameters in parallel with the input, which is more hardware-friendly. Inspired by its efficiency, researchers have extended Mamba architectures to model 2D images and 3D videos on various vision tasks \cite{zhu2024vim, liu2024vmambavisualstatespace, li2024videomamba, he2024mambaad, xing2024segmamba, hu2024zigma, lu2024videomambapro, wang2024pv}. More recently, different Mamba approaches \cite{li2024spikemba, qiao2024vlmamba, he2024panmamba, zhao2024cobra} have been proposed to handle different sequences. VL-Mamba \cite{qiao2024vlmamba} designed a Mamba module to project the features extracted from image and text modalities. Pan-Mamba \cite{he2024panmamba} utilized a shared gate factor for the SSMs from different modalities. However, these methods treat SSMs primarily for intra-sequence modeling and only perform interaction on features after SSMs, 
thereby lacking direct inter-dependency transformations during selective scanning process. This limitation can hinder their ability to fully leverage the complementary information across different inputs. Therefore in this paper, we propose to employ the SSM in enhancing the recurrent feature propagation and reconstruction of video frames.

%% file: sec/3_finalcopy.tex
\section{Methodology}
\label{sec:met}

\subsection{Preliminaries}
State-Space Models (SSMs)~\cite{gu2021efficiently} represent a class of sequence models inspired by control systems, which map a one-dimensional stimulation $x(t) \in \mathbb{R}$ to response $y(t) \in \mathbb{R}$ via a hidden state $h(t) \in \mathbb{R}^N$, which are formulated as linear ordinary differential equations (ODEs):
\vspace{-3pt}
\begin{equation}
\vspace{-3pt}
    \begin{aligned}
        h^{\prime}(t) &= \mathbf{A} h(t)+\mathbf{B} x(t), ~~~ 
        y(t) = \mathbf{C} h(t),
    \end{aligned}
\end{equation}
where the state transition matrix $\mathbf{A} \in \mathbb{R}^{N \times N}$, $\mathbf{B}\in \mathbb{R}^{N \times 1}$ and $\mathbf{C} \in \mathbb{R}^{1 \times N}$ for a state size $N$. To discretize the ODEs, S4~\cite{gu2021efficiently} utilized zero-order hold with a timescale parameter $\Delta$, and transformed the parameters $\mathbf{A}$ and $\mathbf{B}$ from the continuous system into the discrete parameters $\overline {\mathbf{A}}$ and $\overline{\mathbf{B}}$:
\vspace{-3pt}
\begin{equation}
\vspace{-3pt}
    \begin{aligned}
        \overline{\mathbf{A}} &= \exp(\mathbf{\Delta} \mathbf{A}), ~~~
        \overline{\mathbf{B}} = (\mathbf{\Delta A})^{-1}(\exp (\mathbf{\Delta A})-\mathbf{I}) \cdot \mathbf{\Delta B}.
    \end{aligned}
\end{equation}
Therefore, the discretized model can be formulated as:
\vspace{-2pt}
\begin{equation}
\vspace{-2pt}
    \begin{aligned}
        h_i &= \mathbf{\overline{A}}h_{i-1} + \mathbf{\overline{B}}{x}_{i}, ~~~
        y_i = \mathbf{C}h_i.
    \end{aligned}
    \label{eq:ssm_discrete}
\end{equation}

At last, from the perspective of global convolution, the output can be defined as:
\begin{equation}
\begin{aligned}
\mathbf{\overline{K}} &= (\mathbf{C}\mathbf{\overline{B}}, \mathbf{C}\mathbf{\overline{A}}\mathbf{\overline{B}}, \dots, \mathbf{C}\mathbf{\overline{A}}^{L-1}\mathbf{\overline{B}}), ~~~
{y} = \pmb{x} * \mathbf{\overline{K}},
\end{aligned}
\end{equation}
where $*$ represents convolution operation, $L$ is the length of sequence ${x}$, and $\mathbf{\overline{K}} \in \mathbb{R}^L$ is a structured convolutional kernel. This method leverages convolution to synthesize the outputs across the sequence simultaneously, which enhances computational efficiency and scalability.

However, the traditional SSMs like S4 used static parameterization, which restricts the capability to capture the sequence context. To address this, Mamba \cite{gu2023mamba} calculated the dynamic input-dependent parameters $\mathbf{B}$, $\mathbf{C}$, and $\mathbf{\Delta}$ based on the full input sequence ${x}$, allowing a richer, content-aware parameterization and information selection. Then, it projected the full sequence into state-space representations, and selectively scanned over it to learn global features.
Mamba2 \cite{dao2024mamba2} further derived a state-space duality for the dual forms between quadratic-time kernel attention and SSM. It also designed an efficient parallel projection pattern for the $\mathbf{A}$, $\mathbf{B}$, $\mathbf{C}$, ${x}$ inputs by rearranging parameters in chunks. Due to its effectiveness in modeling long-range dependencies and efficiency in time and computation, it holds great potential for medical VSR, which requires accurate detail preservation and fast processing to enhance diagnostic quality.

\begin{figure*}
    \centering
    \includegraphics[width=\textwidth]{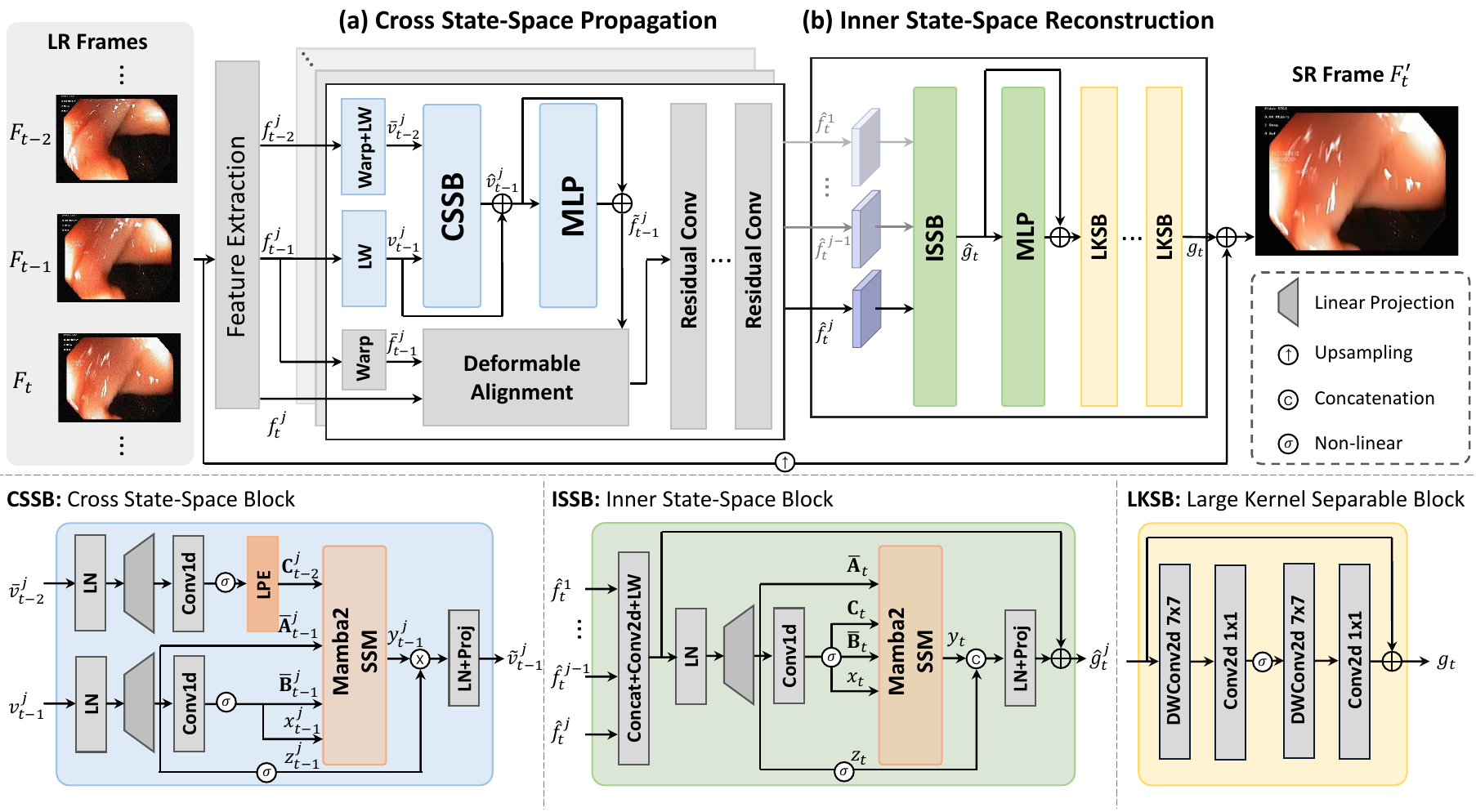}
    \caption{Illustration of the proposed MedVSR framework. For clarity, the main stream shows the $j$-th propagation branch. MedVSR includes two core operations, which are CSSP that captures consistent features for enhancing propagation, and ISSR that reconstructs smooth frames with long-range spatial feature learning and short-range information aggregation.
    }
    \label{fig:main}
    \vspace{-2pt}
\end{figure*}
\subsection{MedVSR Framework Overview}
Given a LR video sequence with frames $F_1, ..., F_T$, our goal is to generate the super-resolved (SR) frames $F_1^{'}, ..., F_T^{'}$ with a specific scaling factor. The overall architecture of our proposed MedVSR is shown in Fig. \ref{fig:main}. It adopts a bidirectional recurrent paradigm where the backward branch propagates features recurrently from the end of the sequence to the beginning. Specifically, it utilizes second-order propagation to incorporate features from the two subsequent frames, thereby refining the feature estimation for the current frame. It is constructed with a feature extraction part with plain convolution, multiple forward and backward Cross State-Space Propagation (CSSP) branches, and an Inner State-Space Reconstruction (ISSR) module to produce the final feature. 

\subsection{Cross State-Space Propagation}

\begin{figure}
    \centering
    \includegraphics[width=0.47\textwidth]{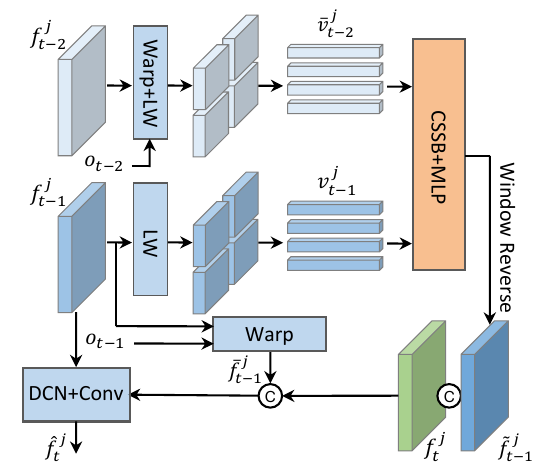}
  \vspace{-6pt}
    \caption{Illustration of the CSSP. It propagates distant frame features to neighbor via the SSM in cross state-space block and aligns with the deformable convolution block. }
    \label{fig:detail_align_module}
  \vspace{-9pt}
\end{figure}

Instead of attempting to utilize the long-distance support frame directly through alignment for propagation, we employ it to control the mapping of the state to the output of the SSM with Cross State-Space Propagation (CSSP), as illustrated in Fig. \ref{fig:main}(a). 
Specifically, let $f_{t-1}^{j}$ represent the feature computed at the $(t-1)$-th timestep in the $j$-th propagation branch. We partition this feature into local window features, denoted as $v_{t-1}^j$. For the feature $f_{t-2}^{j}$ at the $(t-2)$-th frame, we first warp it using the composite flow $o_{t-2} + o_{t-1}$, then also partition it into window features $\overline{v}_{t-2}^j$. The optical flows $o_{t-2}$ and $o_{t-1}$ are estimated using a pretrained SpyNet model \cite{ranjan2017optical} following the common practice \cite{chan2022basicvsr++, liang2022rvrt}.
Then, we treat $\overline{v}_{t-2}^j$ as the supporting feature to support the propagation of feature $v_{t-1}^j$:
\begin{equation}
\begin{aligned}
    \tilde{f}_{t-1}^j &= \text{MLP}(\hat{v}_{t-1}^j) + \hat{v}_{t-1}^j, \\
    \hat{v}_{t-1}^j &= \text{CSSB}(\overline{v}_{t-2}^j, v_{t-1}^j) + {v}_{t-1}^j
\end{aligned}
\end{equation}
where the CSSB and MLP denote the Cross State-Space Block and multi-layer perceptron, respectively. Finally, we utilize $f_{t-1}^j$ to learn the deformable offset for the concatenated feature $[\tilde{f}_{t-1}^j, \overline{f}_{t-1}^j, f_t^j]$ in a DCN layer with residual convolutions \cite{dai2017deformable, wang2019deformable, chan2022basicvsr++}, and obtain the final current frame feature $\hat{f}_t^{j}$. The graphical illustration is presented in Fig. \ref{fig:detail_align_module}. 
%

\textbf{Cross State-Space Block (CSSB).} To facilitate the supporting distant frame to control the propagated feature, 
we need a long-range modeling mechanism
between the distant feature $\overline{v}_{t-2}^j$ and the neighboring feature $v_{t-1}^j$. To achieve better efficiency, we opt for the SSM that is a dual form of the quadratic-time attention \cite{dao2024mamba2} but with significantly less computational cost via a recurrent scanning manner. 
Since the original SSM in Mamba \cite{gu2023mamba, dao2024mamba2} learns the global feature of a single state-space sequence,
we extend it to process two frame sequences with \textit{separate projection layers} followed by 1D convolutions to compute data-dependent parameters: ${x}_{t-1}^j, \overline{\mathbf{B}}_{t-1}^j = \sigma(\text{Conv1d}(\text{LN}(v_{t-1}^j)\mathbf{W}^j))$ and $\mathbf{C}_{t-2}^j = \text{LPE}(\sigma(\text{Conv1d}(\text{LN}(\overline{v}_{t-2}^j) \mathbf{W}_C^j)))$, where LPE represents learnable position embedding that will be discussed next. Notably, $\mathbf{C}_{t-2}^j$ is computed from the state-space sequence of the distant $(t-2)$-th frame. By initializing the hidden state $h^j = {x}_{t-1}^j$ and state matrix $\overline{\mathbf{A}}_{t-1}^j$, the output of each SSM iteration is updated by:
\vspace{-3pt}
\begin{equation}
\vspace{-3pt}
    h_{i}^j = \overline{\mathbf{A}}_{t-1}^j h_{i-1}^j + \overline{\mathbf{B}}_{t-1}^j {x}_{t-1,i}^j, \ 
    {y}_{t-1,i}^j = \mathbf{C}_{t-2}^j h_{i}^j.
\end{equation}
Therefore, the propagated feature ${y}_{t-1}^j$ is a the hidden space of the $(t-1)$-th frame controlled by the $(t-2)$-th frame.
For the next $(j+1)$-th propagation branch, CSSB will also use the control matrix $\mathbf{C}_{t-2}^{j+1}$ derived from $\overline{v}_{t-2}^{j+1}$ to generate the output ${y}_{i+1}^{j+1}$. Progressively, CSSB can compound the state-space sequences across frames and learn to propagate relevant features. After finishing the scan of entire sequences, we utilize the gate function $z_{t-1}^j = \sigma((\text{LN}(v_{t-1}^j)\mathbf{W}^j))$ and the scanned output $y_{t-1}^j$ to produce the propagated output $\tilde{v}_{t-1}^j = \text{LN}(z_{t-1}^j \times y_{t-1}^j)\mathbf{W}^j_{o}$. 
%
%

\textbf{Local Windows (LW).} 
Different from 2D convolution that aggregates local visual features, Mamba scans over all locations recurrently, which could suffer from forgetting of early pixels \cite{guo2024mambair, yuan2024remamba}. This is particularly pronounced in VSR scenarios where the spatial resolution is much larger.
Moreover, tokens within a local area are sufficient in providing critical support for effective reconstruction \cite{liu2021swin, chen2023activating, ray2024cfat, liu2023efficientvit}. 
To this end, we introduce a local window partitioning process prior to CSSB, as shown in Fig. \ref{fig:detail_align_module}. 
For an input feature with dimension $H\times W\times C$, we define a window of size $l$ and partition the feature into $(H/l) \times (W/l)$ local windows, each with a dimension of $ l \times l \times C$. This strategy allows for a more focused analysis of local structures within the data.

\textbf{Learnable Position Embedding (LPE).} In the original Mamba \cite{gu2023mamba}, there is no explicit position embedding \cite{Dosovitskiy2021ViT} as the state-space model operates in a recurrent way. Consequently, the hidden state captures positional information based solely on the incoming order of tokens. However, flattening 2D frames into 1D sequences causes the loss of information regarding the beginning of new rows, hindering the model's ability to identify spatial relations in arbitrary directions. 
Therefore, we propose to employ a LPE layer that incorporates absolute position information and spatial context. Specifically, we reshape the 1D sequence to 2D feature and feed it into a learnable zero-padded 2D depthwise convolution \cite{chollet2017xception}, then reshape it back to 1D and proceed to the SSM. The LPE layer could incorporate absolute position \cite{kayhan2020translation} and improve the learning of spatial relations, which is essential for feature propagation and reconstruction.

\subsection{Inner State-Space Reconstruction}

After obtaining the features propagated from multiple forward and backward CSSP branches, we feed all features $\hat{f}_t^{1:j}$ into the proposed Inner State-Space Reconstruction (ISSR) for reconstruction, which is illustrated in Fig. \ref{fig:main}(b). It uses ISSB with MLP that facilitates long-range spatial feature learning, followed by stacked LKSB with large-kernel convolutions for local information aggregation, leading to a detailed reconstructed feature of the $t$-th frame $g_t$.

\textbf{Inner State-Space Block (ISSB).} Given propagated features from different branches, we first perform a channel-wise concatenation and dimension reduction with a 2D convolution, then split them into local windows:
\begin{equation}
    \hat{v}_t = \text{LW}(\text{Conv2d}([\hat{f}_t^1, ..., \hat{f}_t^{j-1}, \hat{f}_t^j])).
\end{equation}
Considering the large spatial size of the concatenated feature, SSM can be effective utilized for modeling the long-range spatial dependencies. We project the feature into data-dependent parameters: ${x}_t, \overline{\mathbf{B}}_t, \mathbf{C}_{t} = \sigma(\text{Conv1d}(\text{LN}( \hat{v}_{t})\mathbf{W}))$, $z_t = \sigma(\text{LN}(\hat{v}_t)\mathbf{W})$ and apply SSM \cite{dao2024mamba2} following Eq. (\ref{eq:ssm_discrete}) to produce scanned result $y_t$. Specially, the gating mechanism is replaced by a concatenation to produce the output:
\vspace{-3pt}
\begin{equation}
\vspace{-3pt}
    \hat{g}_t = \text{LN}([y_t, \  z_t])\mathbf{W}_o + \hat{v}_t.
\end{equation}
The concatenation operation enhances the learned representation with less computational complexity, which preserves more fine-grained features for reconstruction.

\textbf{Large Kernel Separable Block (LKSB).} Besides learning long-range features via SSMs, we propose to reconstruct the features by gathering short-range information with large kernel convolutions:
\vspace{-4pt}
\begin{equation}
\vspace{-4pt}
    g_t = \text{LKSB}(\text{MLP}(\hat{g}_t) + \hat{g}_t).
\end{equation}
Instead of stacking small kernel convolutions \cite{wang2019edvr, chan2021basicvsr, chan2022basicvsr++}, utilizing large kernels could capture both contextual information from distant regions and details from nearby pixels \cite{liu2022convnet, ding2022scaling, wang2023internimage}, thereby improving the overall smoothness and fidelity of the reconstructed frames.

%% file: sec/4_experiments.tex
\section{Experiments}
\label{sec:exp}

\begin{table*}[htbp]
  \centering
  \caption{MedVSR performance on HyperKvasir \cite{Borgli2020HyperKvasir}, LDPolyp \cite{ma2021ldpolypvideo}, and EndoVis18 \cite{allan2020endovis18} datasets with comparisons to state-of-the-art models.
  All models are trained from scratch with identical settings. The FLOPs and Latency are tested on Nvidia 4090 GPU using the video clip in HyperKvasir testing dataset. Best performances are highlighted in bold and second best performances are underlined.}
  \vspace{-4pt}
\setlength\tabcolsep{7.3pt}
    \scalebox{1.02}{\begin{tabular}{l|ccc|cccccc}
    \toprule
    \multirow{2}[2]{*}{Model} & \multirow{1}[2]{*}{Params}  & \multirow{1}[2]{*}{FLOPs} & \multirow{1}[2]{*}{Latency} & \multicolumn{2}{c}{HyperKvasir \cite{Borgli2020HyperKvasir}}  & \multicolumn{2}{c}{LDPolyp \cite{ma2021ldpolypvideo}} & \multicolumn{2}{c}{EndoVis18 \cite{allan2020endovis18}} \\
\cmidrule{5-10}          & (M) & (T) & (s) & PSNR   & SSIM   & PSNR   & SSIM  & PSNR   & SSIM  \\
\midrule
    EDVR \cite{wang2019edvr} & 20.63& 45.41 & 2.4398 & 27.1180 & 0.8483 & 30.0810&0.8417 & 21.6726 & 0.7883 \\
    BasicVSR \cite{chan2021basicvsr} &6.29&8.56 & 0.5018 &31.4643&0.8990& 31.6828&0.8650 & 30.7162&0.8950\\
    BasicVSR++ \cite{chan2022basicvsr++} &7.32& 9.47 & 0.7547 &31.7270& \underline{0.9042}& 31.6685 & 0.8620 & \underline{30.7977} & \underline{0.8958} \\ 
    VSRT \cite{cao2021vsrt} &32.56 & 112.64 & 8.1197 & \underline{31.7789} & 0.8999 &31.6287&0.8596 & 30.5871 & 0.8916 \\
    RVRT \cite{liang2022rvrt} &10.78 & 44.36  & 16.0213 &  29.2621 & 0.8944 & 31.3174&0.8578 & 30.7222 & 0.8932 \\
    TCNet \cite{liu2022tcnet} &9.64 & 39.55 & 9.1565 &31.1137 & 0.8889& 31.3888&0.8547 & 30.1653 & 0.8859 \\
    IART \cite{xu2024iart} & 13.41 &44.71&21.7553& 31.2984 & 0.9030 & \underline{31.6994} & \underline{0.8671} & 30.4658 & 0.8924 \\
    \rowcolor[rgb]{ .949,  .949,  .949} 
    MedVSR (Ours) &7.16 & 9.46 & 1.1486 & \textbf{32.0958}& \textbf{0.9069}& \textbf{31.8333} & \textbf{0.8673} & \textbf{30.8344}& \textbf{0.8960} \\

    \bottomrule
    \end{tabular}}%
  \label{tab:hk_v4}%
  \vspace{-6pt}
\end{table*}%

\begin{figure*}
    \centering
    \includegraphics[width=0.99\linewidth]{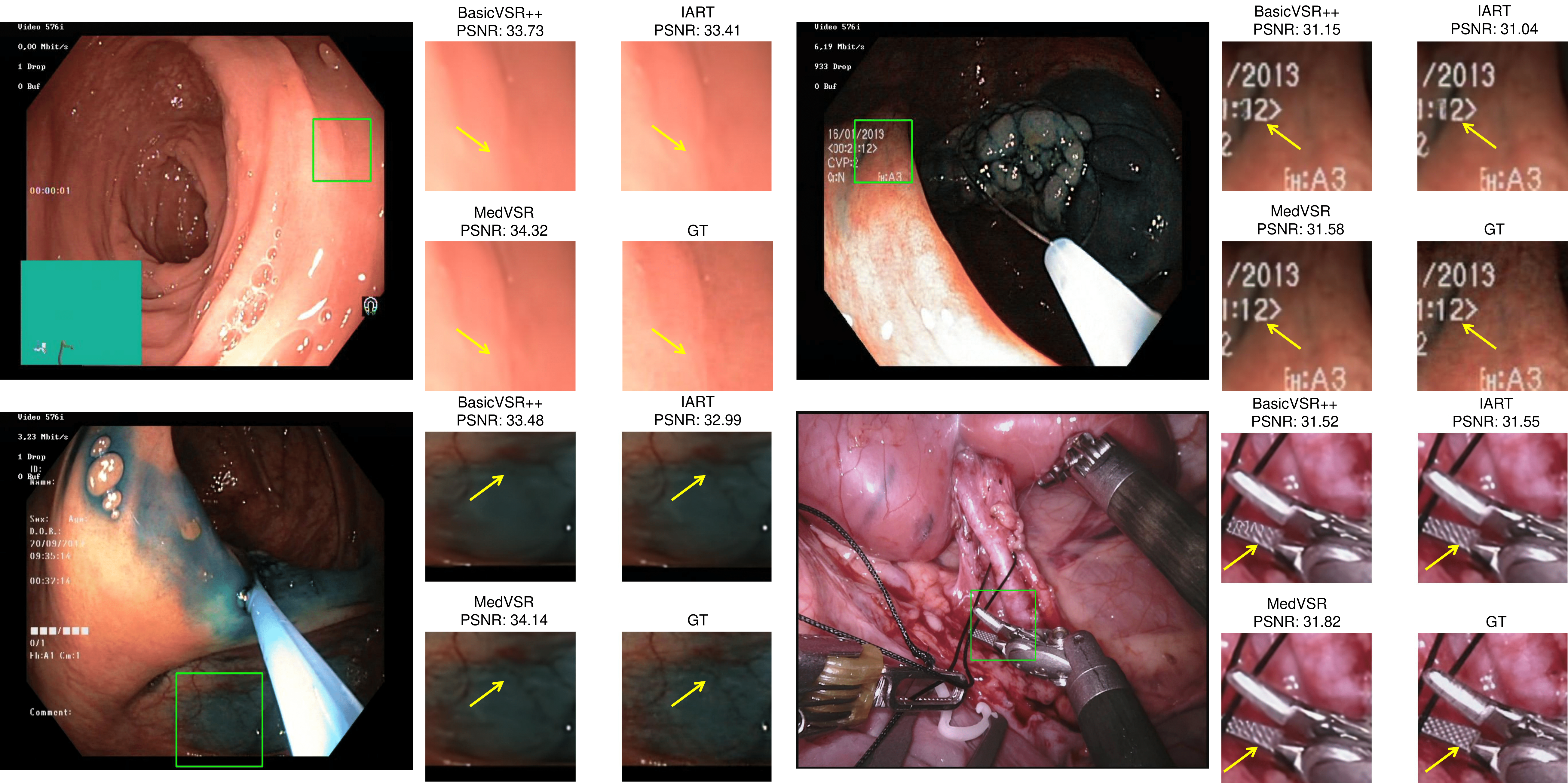}
    \caption{Qualitative comparisons on HyperKvasir \cite{Borgli2020HyperKvasir}, LDPolyp \cite{ma2021ldpolypvideo}, and EndoVis18 \cite{allan2020endovis18}. MedVSR reduces artifacts, provides detailed results, and reconstructs the textures accurately. For more qualitative comparisons, please refer to supplementary. \textit{Zoom in for details.}}
    \label{fig:vis_comparison}
  \vspace{-6pt}
\end{figure*}
\subsection{Datasets and Implementation Details} 
%
To evaluate our MedVSR, we conduct experiments on several medical video datasets with distinct scenes and screening devices, including HyperKvasir \cite{Borgli2020HyperKvasir}, LDPolyp \cite{ma2021ldpolypvideo}, EndoVis18 \cite{allan2020endovis18}, and Cataract-101 \cite{cataract}. The HyperKvasir dataset is divided into training, validation, and testing sets. The Cataract-101 dataset is divided into training and testing sets. We train all models for 100,000 iterations, with a learning rate of 2e-4 and a cosine learning rate decay with a minimum value of 1e-7. The degradation process involves 4$\times$ bicubic downsampling accompanied by Gaussian noise with a covariance of 15. 
The number of propagation branches $j$ is 4 in our framework.
During training, HR patches of size 
256$\times$256 are used. For the models trained with HyperKvasir training set, we test them on HyperKvasir testing set, LDPolyp, and EndoVis18, as they are all capturing the intra-body structures. The models trained with Cataract-101 training set are evaluated on its corresponding testing set.
We follow existing VSR works \cite{chan2021basicvsr, xu2024iart} to train the model with Charbonnier loss  \cite{charbonnier1994two}
$\mathcal{L} = \sqrt{|| F'_t - I_{t} || ^{2} + \epsilon^{2}}$, with $F'_t$ as the reconstructed frame and $I_t$ as the ground truth frame. Peak Signal-to-Noise Ratio (PSNR) and Structural Similarity Index Measure (SSIM) are used as evaluation metrics.

\subsection{Comparison with State-of-the-Art Methods}

We compare our MedVSR with several state-of-the-art models, including EDVR \cite{wang2019edvr}, BasicVSR \cite{chan2021basicvsr}, BasicVSR++ \cite{chan2022basicvsr++}, VSRT \cite{cao2021vsrt}, RVRT \cite{liang2022rvrt}, TCNet \cite{liu2022tcnet}, and IART \cite{xu2024iart}. These models are trained from scratch with identical settings to ensure a fair comparison. 
As shown in Tab.~\ref{tab:hk_v4}, our MedVSR model outperforms compared models on all datasets.
On the HyperKvasir dataset, MedVSR achieves a PSNR of 32.0958, which is 0.3169  higher than the second-best model VSRT \cite{cao2021vsrt}, while requiring 3.5$\times$ fewer parameters and 10.9$\times$ fewer FLOPs. 
Similarly, on the LDPolyp dataset, MedVSR achieves a PSNR of 31.8333 and an SSIM of 0.8673, which are both higher than IART \cite{xu2024iart}, while using only 3.6$\times$ fewer FLOPs and 3.7$\times$ fewer parameters. 
Notably, in the surgical scene dataset EndoVis18, which exhibits a larger gap and frame difference, our MedVSR remains effective, achieving a PSNR of 30.8344 and an SSIM of 0.8960. While the improvement over the powerful BasicVSR++ \cite{chan2022basicvsr++} may not be significant in terms of metrics, it excels in capturing nuanced details in the final super-resolved frames.
This suggests that our model is also well-suited for handling complex and dynamic surgical scenes.

In addition to its better performance, MedVSR model also demonstrates a significant advantage in inference speed, outpacing recent transformer-based models remarkably \cite{cao2021vsrt, liang2022rvrt, xu2024iart}, while achieving comparable speed to CNN-based models that have been highly optimized by modern deep learning hardware \cite{chan2021basicvsr, chan2022basicvsr++, liu2022tcnet}. 
Specifically, when measured on the HyperKvasir testing dataset with 50 frames per clip, the inference latency of MedVSR is 1.1642s, which is $6.0\times$ faster than VSRT, $12.7\times$ faster than RVRT, and $18.7\times$ faster than IART.
The results demonstrate the efficacy and efficiency of using state-space transformations for propagation and reconstruction, establishing it as a robust approach for the complex and demanding medical VSR task.

We present a qualitative comparison in Fig.~\ref{fig:vis_comparison}. BasicVSR++ and IART could suffer from artifacts and blurry textures, while MedVSR successfully reconstructs the subtle vessels. Besides, MedVSR can restore the text precisely, which could be more clinically informative. The bottom right example shows that MedVSR preserves the texture of surgical tools. Notably, super-resolved tool textures offer several benefits for doctors. They can {reveal micro-scratches} and blade deformities, monitoring {tool status} and preventing tool failure. Also, sharper textures could {enhance the accuracy of automatic tool control systems} by providing more detailed visual cues.

\begin{figure}
    \centering
    \includegraphics[width=0.99\linewidth]{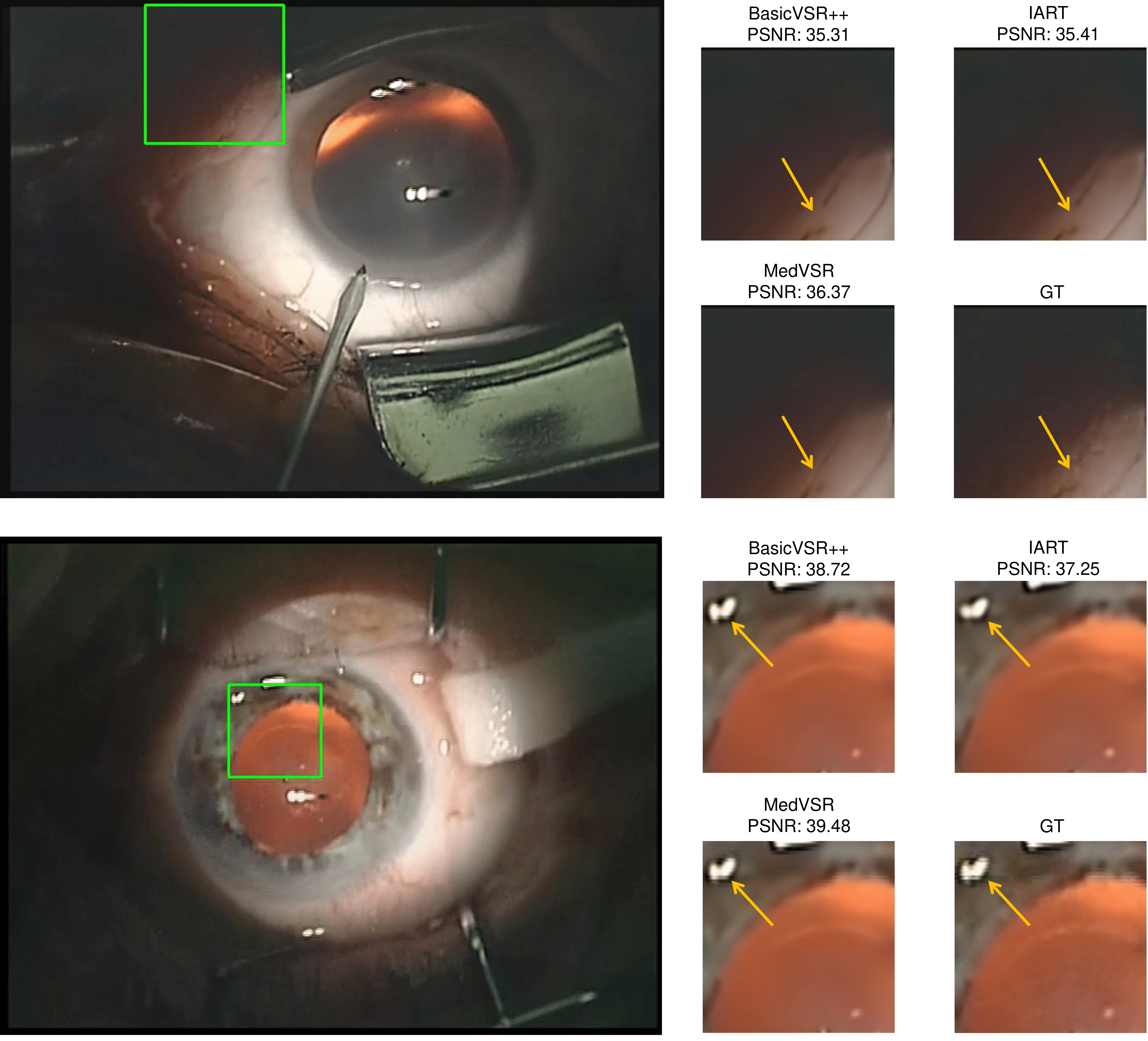}
  \vspace{-4pt}
    \caption{Qualitative comparisons on Cataract-101 \cite{cataract}.}
  \vspace{-12pt}
    \label{fig:vis_comparison_cat}
\end{figure}

In Tab.~\ref{tab:cartar}, we list the results on the Cataract-101 dataset, which comprises different operation phases during cataract surgeries. MedVSR remains effective in different medical scenarios, which achieves 36.2275 PSNR and 0.9308 SSIM. Compared to TCNet \cite{liu2022tcnet} that explores long-distance consistency, our model surpasses it by 1.6280 in PSNR and 0.0131 in SSIM, meanwhile using $0.3\times$ fewer parameters and $3.4\times$ fewer FLOPs. In Fig.~\ref{fig:vis_comparison_cat}, qualitative results are provided. From the top case, both BasicVSR++ \cite{chan2022basicvsr++} and IART \cite{xu2024iart} produce results with inconsistent edges, which do not reflect the authentic structure. In contrast, MedVSR reconstructs the full structure precisely. In the bottom case, MedVSR preserves more detailed structure and color of the original frame. The results further demonstrate the generalization ability and robustness of our MedVSR.

\begin{table}[t]
\vspace{-6pt}
  \centering
  \caption{MedVSR performance on Cataract-101 \cite{cataract}. FLOPs is tested on NVIDIA 4090 GPU using the video clip in Cataract-101 test dataset.
  Best performances are highlighted in bold and second best performances are underlined.}
    \scalebox{1.02}{\begin{tabular}{l|cc|cc}
    \toprule
    \multirow{2}[2]{*}{Model} & \multirow{1}[2]{*}{Params} & \multirow{1}[2]{*}{FLOPs} & \multicolumn{2}{c}{Cataract-101 \cite{cataract}}   \\
\cmidrule{4-5}        & (M) & (T)  & PSNR   & SSIM    \\
\midrule
    EDVR \cite{wang2019edvr} & 20.63&42.89& 31.7843 & 0.9103\\
    BVSR \cite{chan2021basicvsr} &6.29&8.09&35.2892 & 0.9232\\
    BVSR++ \cite{chan2022basicvsr++} &7.32&8.94 &35.5358 & 0.9251\\ 
    VSRT \cite{cao2021vsrt} &32.56 &112.64& 35.2495 & 0.9230\\
    RVRT \cite{liang2022rvrt} &10.78 &44.36& 35.6378 & 0.9281 \\
    TCNet \cite{liu2022tcnet} &9.64 &39.55&34.5995 & 0.9177\\
    IART \cite{xu2024iart} & 13.41 &44.71& 36.0764 & 0.9275 \\
    \rowcolor[rgb]{ .949,  .949,  .949} MedVSR &7.16 &8.95& \textbf{36.2275} & \textbf{0.9308} \\
    \bottomrule
    \end{tabular}}%
  \label{tab:cartar}%
  \vspace{-6pt}
\end{table}%

\subsection{Ablation Study}
We conduct ablation studies on the HyperKvasir dataset \cite{Borgli2020HyperKvasir} from the following aspects.

\textbf{Impact of the components in CSSB.} We ablate CSSB by removing specific elements: the learnable position embedding (LPE), the local windows (LW), and the separate projection layers (SP). As shown in Tab. \ref{tab:ablation_cssb}, the CSSB without LPE shows a 0.2856 PSNR degradation, demonstrating the efficacy of incorporating absolute position information for distant frames. Without LW, the PSNR drops by 0.2757, which indicates that scanning within local regions enhances the details more effectively. Lastly, the CSSB without SP is similar to directly concatenating the features of the two frames and process them jointly. This yields a large PSNR drop by 0.3572, which indicates the large visual gap in distant frames is difficult for the simple optical flow-based methods to propagate. The result highlights the importance of the learned consistent and stable information.

\textbf{Impact of the components in ISSB.} We assess the design of ISSB, including local windows (LW) and the concatenation (CAT) in Tab. \ref{tab:ablation_issb}. 
Firstly, the model without LW shows a decrease in PSNR by 0.2743, suggesting that the information from early pixels during recurrent scanning are forget thus is ineffective.
Secondly, replacing the original gating operation with CAT can reduce the parameters by 0.01M while improving the performance by 0.2430.

\textbf{Impact of the LKSB design.} We propose LKSB to enhance the short-range information aggregation during reconstruction while alleviates artifacts, and we compare our design with other common convolution blocks in Tab. \ref{tab:ablation_lksb}. Our model surpasses residual block \cite{resnet} and depth-wise seperable block \cite{sandler2018mobilenetv2} remarkably using 0.04M and 0.08M fewer parameters. Another choice is the partial block \cite{chen2023run} that was proposed to improve efficiency, but it leads to a large decline in PSNR by 0.2152. Moreover, we ablate the kernel size. The performance continuous improves as the kernel size increases and saturates at 7. When increasing it to 9, the PSNR drops by 0.2106, which suggests that the short-range information cannot be effectively aggregated. Therefore, we set the kernel size to 7 that achieves the optimal computational cost and performance trade-off. 

\begin{table}[t]
  \centering
  \caption{Ablation study on the components in CSSB.}
  \vspace{-6pt}
\scalebox{1.00}{\begin{tabular}{l|cccc}
\toprule
Ablation       & Params  & FLOPs & PSNR & SSIM \\ \midrule
CSSB $_{\text{w/o LPE}}$ & 7.16       &   9.46    & 31.8102     &  0.9049    \\
CSSB $_{\text{w/o LW}}$    & 7.16       &  9.45      & 31.8201     & 0.9050     \\
CSSB $_{\text{w/o SP}}$    &   7.15     &  9.44     & 31.7386    & 0.9047    \\
\rowcolor[rgb]{ .949,  .949,  .949}CSSB           & 7.16       &  9.46     &  32.0958    & 0.9069    \\ \bottomrule
\end{tabular}}
\label{tab:ablation_cssb}
  \vspace{-4pt}
\end{table}

\begin{table}[]
  \centering
  \caption{Ablation study on the components in ISSB.}
  \vspace{-6pt}
\scalebox{1.00}{\begin{tabular}{l|cccc}
\toprule
Ablation       & Params & FLOPs& PSNR & SSIM \\ \midrule
ISSB $_{\text{w/o LW}}$ & 7.16       & 9.45      &   31.8215   & 0.9053     \\
ISSB $_{\text{w/o CAT}}$    & 7.17       & 9.46      & 31.8528     & 0.9055     \\
\rowcolor[rgb]{ .949,  .949,  .949}ISSB           & 7.16       &  9.46     &  32.0958    & 0.9069   \\ \bottomrule
\end{tabular}}
\label{tab:ablation_issb}
  \vspace{-4pt}
\end{table}

\begin{table}[]
  \centering
  \caption{Ablation study on the design of LKSB.}
  \vspace{-6pt}
\scalebox{0.97}{\begin{tabular}{l|cccc}
\toprule
Ablation       & Params & FLOPs& PSNR & SSIM \\ \midrule
ResBlock~\cite{resnet}   &  7.20      &   9.50    & 31.8067     & 0.9053     \\
DWBlock~\cite{sandler2018mobilenetv2}   & 7.24       &  9.51     &  31.7858    &  0.9003    \\
PBlock~\cite{chen2023run}    &  7.12      & 9.41      & 31.8806     & 0.9041     \\
3$\times$3 LKSB           &    7.12    &  9.40     & 31.9049     & 0.9050     \\ 
5$\times$5 LKSB           &    7.13    & 9.42      & 31.9077 & 0.9052 \\ 
\rowcolor[rgb]{ .949,  .949,  .949}7$\times$7 LKSB           &    7.16    & 9.46      &  32.0958    & 0.9069     \\ 
9$\times$9 LKSB           &    7.20    &  9.51     &    31.8852  & 0.9037     \\ \bottomrule
\end{tabular}}
\label{tab:ablation_lksb}
  \vspace{-10pt}
\end{table}
\textbf{Impact of the propagation scheme.} 
In the proposed CSSB, we utilize ($t\scalebox{0.75}[1.0]{\( - \)}2$)-th frame to support ($t\scalebox{0.75}[1.0]{\( - \)}1$)-th frame via SSM, which could extract stable and continuous features across frames. Here we ablate other propagation schemes in Tab. \ref{tab:ablation_prop_scheme}. When using ($t\scalebox{0.75}[1.0]{\( - \)}2$)-th frame to support the $t$-th frame ($t\scalebox{0.75}[1.0]{\( - \)}2 \scalebox{0.9}[1.0]{\( \rightarrow \)} t$), the performance drops slightly as the captured consistent representations are not sufficient. 
When propagating the feature only from $t\scalebox{0.75}[1.0]{\( - \)}1 \scalebox{0.9}[1.0]{\( \rightarrow \)} t$, PSNR drops by 0.3224, which shows that without the long-term consistency captured at the ($t\scalebox{0.75}[1.0]{\( - \)}2$)-th frame, the model fails to leverage the temporal information effectively. 
Moreover, when combining the $t\scalebox{0.75}[1.0]{\( - \)}2 \scalebox{0.9}[1.0]{\( \rightarrow \)} t$ and $t\scalebox{0.75}[1.0]{\( - \)}1 \scalebox{0.9}[1.0]{\( \rightarrow \)} t$ (Both), the performance is improved to 31.9116 PSNR, yet remains  inferior to our design using $t\scalebox{0.75}[1.0]{\( - \)}2 \scalebox{0.9}[1.0]{\( \rightarrow \)} t\scalebox{0.75}[1.0]{\( - \)}1$ with 32.0958 PSNR. The result suggests that the $t$-th frame feature is already well extracted and may not be modified, while CSSB could produce well-aligned propagated features to enhance representations.

\subsection{Further Analysis}

\begin{figure}
    \centering
    \includegraphics[width=0.89\linewidth]{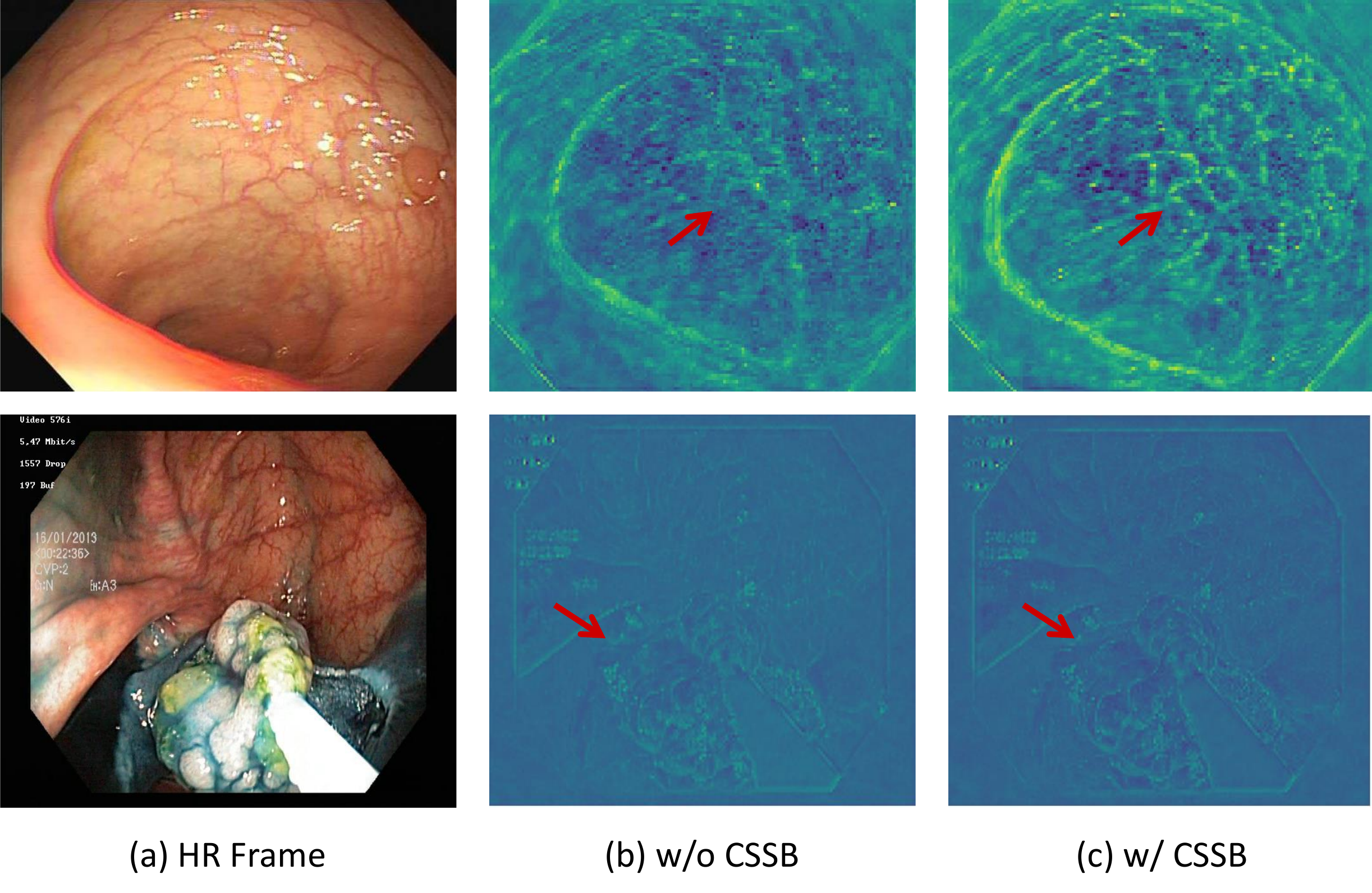}
  \vspace{-4pt}
    \caption{Visualization of the effect of Cross State-Space Block. (a) The corresponding HR frame. (b) The propagated feature without the proposed CSSB. (c) The propagated feature with the proposed CSSB. \textit{Best viewed in color.}}
    \label{fig:feat_comparison}
  \vspace{-12pt}
\end{figure}

\textbf{Visualization of the effect of CSSB.} In Fig. \ref{fig:feat_comparison}, we visualize the effect of utilizing our proposed CSSB. As shown in Fig. \ref{fig:feat_comparison}(b), the propagated features $\hat{f}_{t}^j$ without CSSB could lack the necessary details and lead to the presence of artifacts in the medical videos. Conversely, we observe a significant enhancement in the clarity and of the visualized tissues and vessels with CSSB in Fig. \ref{fig:feat_comparison}(c). CSSB not only captures finer details but also reduces noise, which provides a more accurate representation of the underlying anatomy. 

\textbf{Analysis on the window size in LW.} We investigate the impact of local window size $l$ on our model's performance. As shown in Tab. \ref{tab:ablation_window_size}, the best performance is achieved with a window size of 16, while a window size of 8 yields competitive results. A larger window size of 32 leads to a degradation in performance. This suggests that a moderate size 16 is optimal for capturing long-range spatial features.

\begin{table}[]
  \centering
  \caption{Ablation study on the propagation scheme in CSSB.}
  \vspace{-6pt}
\scalebox{1.00}{\begin{tabular}{l|ccac}
\toprule
Metric       &
$t\scalebox{0.75}[1.0]{\( - \)}2 \scalebox{0.9}[1.0]{\( \rightarrow \)} t$ 
&
$t\scalebox{0.75}[1.0]{\( - \)}1 \scalebox{0.9}[1.0]{\( \rightarrow \)} t$ &
$t\scalebox{0.75}[1.0]{\( - \)}2 \scalebox{0.9}[1.0]{\( \rightarrow \)} t\scalebox{0.75}[1.0]{\( - \)}1$ 
& Both \\ \midrule
PSNR          & 31.8684       &  31.7734     &  32.0958    & 31.9116   \\ 
SSIM          & 0.9053       &  0.9044     &  0.9069    & 0.9046   \\ 
\bottomrule
\end{tabular}}
\label{tab:ablation_prop_scheme}
  \vspace{-5pt}
\end{table}

\begin{table}[]
  \centering
  \caption{Performance comparison with different window sizes.}
  \vspace{-6pt}
\scalebox{1.00}{\begin{tabular}{l|ccac}
\toprule
Metric       &
$l=\ $4 & $l=\ $8 & $l=\ $16 & $l=\ $32 \\ \midrule
PSNR          & 31.8661       &  31.9016     &  32.0958    & 31.8366   \\ 
SSIM          & 0.9051       &  0.9051     &  0.9069    & 0.9048   \\ 
\bottomrule
\end{tabular}}
  \vspace{-5pt}
\label{tab:ablation_window_size}
\end{table}

%% file: sec/5_conclusion.tex
\vspace{-3pt}
\section{Conclusion}
\label{sec:con}
\vspace{-3pt}

In this paper, we point out that medical videos contain abrupt frame transitions and jitter, leading to optical flow errors and alignment difficulties. Besides, the inherent continuous tissue structures and nuanced vessels present difficulties for existing VSR models, which tend to reconstruct distorted features and are prone to artifacts. To this end, we propose MedVSR, which contains a CSSP module that utilizes distant frames to guide the propagation process and capture consistent features, and an ISSR module that jointly learns long-range spatial features and aggregates local information. Extensive experimental results demonstrate the superior performance and efficiency of MedVSR.

%% file: main.bib
@String(ICLR = {Int. Conf. Learn. Represent.})

@String(AAAI = {AAAI})

@String(ICLR  = {ICLR})

@article{Borgli2020HyperKvasir,
  title = {{HyperKvasir, a comprehensive multi-class
    image and video dataset for gastrointestinal endoscopy}},
  author = {
    Borgli, Hanna and Thambawita, Vajira and
    Smedsrud, Pia H and Hicks, Steven and Jha, Debesh and
    Eskeland, Sigrun L and Randel, Kristin Ranheim and
    Pogorelov, Konstantin and Lux, Mathias and
    Nguyen, Duc Tien Dang and Johansen, Dag and
    Griwodz, Carsten and Stensland, H{\aa}kon K and
    Garcia-Ceja, Enrique and Schmidt, Peter T and
    Hammer, Hugo L and Riegler, Michael A and
    Halvorsen, P{\aa}l and de Lange, Thomas
  },
  doi = {10.1038/s41597-020-00622-y},
  issn = {2052-4463},
  journal = {Scientific Data},
  number = {1},
  pages = {283},
  url = {https://doi.org/10.1038/s41597-020-00622-y},
  volume = {7},
  year = {2020}
}

@inproceedings{wang2024pv,
  title={PV-SSM: Exploring Pure Visual State Space Model for High-dimensional Medical Data Analysis},
  author={Wang, Cheng and Liu, Xinyu and Li, Chenxin and Liu, Yifan and Yuan, Yixuan},
  booktitle={2024 IEEE International Conference on Bioinformatics and Biomedicine (BIBM)},
  pages={2542--2549},
  year={2024},
  organization={IEEE}
}

@article{liu2025endogen,
  title={EndoGen: Conditional Autoregressive Endoscopic Video Generation},
  author={Liu, Xinyu and Liu, Hengyu and Wang, Cheng and Liu, Tianming and Yuan, Yixuan},
  journal={arXiv preprint arXiv:2507.17388},
  year={2025}
}

@inproceedings{liu2023efficientvit,
  title={Efficientvit: Memory efficient vision transformer with cascaded group attention},
  author={Liu, Xinyu and Peng, Houwen and Zheng, Ningxin and Yang, Yuqing and Hu, Han and Yuan, Yixuan},
  booktitle={Proceedings of the IEEE/CVF conference on computer vision and pattern recognition},
  pages={14420--14430},
  year={2023}
}

@inproceedings{zhu2024vim,
  title={Vision Mamba: Efficient Visual Representation Learning with Bidirectional State Space Model},
  author={Zhu, Lianghui and Liao, Bencheng and Zhang, Qian and Wang, Xinlong and Liu, Wenyu and Wang, Xinggang},
  booktitle={International Conference on Machine Learning}
}

@inproceedings{kayhan2020translation,
  title={On translation invariance in cnns: Convolutional layers can exploit absolute spatial location},
  author={Kayhan, Osman Semih and Gemert, Jan C van},
  booktitle={Proceedings of the IEEE/CVF Conference on Computer Vision and Pattern Recognition},
  pages={14274--14285},
  year={2020}
}

@inproceedings{xing2024segmamba,
  title={Segmamba: Long-range sequential modeling mamba for 3d medical image segmentation},
  author={Xing, Zhaohu and Ye, Tian and Yang, Yijun and Liu, Guang and Zhu, Lei},
  booktitle={International Conference on Medical Image Computing and Computer-Assisted Intervention},
  pages={578--588},
  year={2024},
  organization={Springer}
}

@article{he2024mambaad,
  title={Mambaad: Exploring state space models for multi-class unsupervised anomaly detection},
  author={He, Haoyang and Bai, Yuhu and Zhang, Jiangning and He, Qingdong and Chen, Hongxu and Gan, Zhenye and Wang, Chengjie and Li, Xiangtai and Tian, Guanzhong and Xie, Lei},
  journal={arXiv preprint arXiv:2404.06564},
  year={2024}
}

@article{hu2024zigma,
  title={Zigma: Zigzag mamba diffusion model},
  author={Hu, Vincent Tao and Baumann, Stefan Andreas and Gui, Ming and Grebenkova, Olga and Ma, Pingchuan and Fischer, Johannes and Ommer, Bjorn},
  journal={arXiv preprint arXiv:2403.13802},
  year={2024}
}

@article{li2024videomamba,
  title={Videomamba: State space model for efficient video understanding},
  author={Li, Kunchang and Li, Xinhao and Wang, Yi and He, Yinan and Wang, Yali and Wang, Limin and Qiao, Yu},
  journal={arXiv preprint arXiv:2403.06977},
  year={2024}
}

@inproceedings{sandler2018mobilenetv2,
  title={Mobilenetv2: Inverted residuals and linear bottlenecks},
  author={Sandler, Mark and Howard, Andrew and Zhu, Menglong and Zhmoginov, Andrey and Chen, Liang-Chieh},
  booktitle={Proceedings of the IEEE Conference on Computer Vision and Pattern Recognition},
  pages={4510--4520},
  year={2018}
}

@inproceedings{chen2023activating,
  title={Activating more pixels in image super-resolution transformer},
  author={Chen, Xiangyu and Wang, Xintao and Zhou, Jiantao and Qiao, Yu and Dong, Chao},
  booktitle={Proceedings of the IEEE/CVF Conference on Computer Vision and Pattern Recognition},
  pages={22367--22377},
  year={2023}
}

@inproceedings{ray2024cfat,
  title={CFAT: Unleashing Triangular Windows for Image Super-resolution},
  author={Ray, Abhisek and Kumar, Gaurav and Kolekar, Maheshkumar H},
  booktitle={Proceedings of the IEEE/CVF Conference on Computer Vision and Pattern Recognition},
  pages={26120--26129},
  year={2024}
}

@article{almalioglu2020endol2h,
  title={EndoL2H: deep super-resolution for capsule endoscopy},
  author={Almalioglu, Yasin and Ozyoruk, Kutsev Bengisu and Gokce, Abdulkadir and Incetan, Kagan and Gokceler, Guliz Irem and Simsek, Muhammed Ali and Ararat, Kivanc and Chen, Richard J and Durr, Nicholas J and Mahmood, Faisal and others},
  journal={IEEE Transactions on Medical Imaging},
  volume={39},
  number={12},
  pages={4297--4309},
  year={2020},
  publisher={IEEE}
}

@inproceedings{hafner2013pocs,
  title={POCS-based super-resolution for HD endoscopy video frames},
  author={H{\"a}fner, Michael and Liedlgruber, Michael and Uhl, Andreas},
  booktitle={Proceedings of the 26th IEEE International Symposium on Computer-Based Medical Systems},
  pages={185--190},
  year={2013},
  organization={IEEE}
}

@article{hayat2024sevsr,
  title={E-SEVSR-Edge Guided Stereo Endoscopic Video Super-Resolution},
  author={Hayat, Mansoor and Aramvith, Supavadee},
  journal={IEEE Access},
  year={2024},
  publisher={IEEE}
}

@article{song2022deformable,
  title={Deformable transformer for endoscopic video super-resolution},
  author={Song, Xiaowei and Tang, Hui and Yang, Chunfeng and Zhou, Guangquan and Wang, Yangang and Huang, Xinjun and Hua, Jie and Coatrieux, Gouenou and He, Xiaopu and Chen, Yang},
  journal={Biomedical Signal Processing and Control},
  volume={77},
  pages={103827},
  year={2022},
  publisher={Elsevier}
}

@article{zhou2022video,
  title={Video super-resolution for wireless capsule endoscopy imaging sensor},
  author={Zhou, Chao and Qiu, Kunpeng and Chen, Can and Zhang, Dengyin and Guo, Yongxin},
  journal={IEEE Sensors Journal},
  volume={22},
  number={17},
  pages={17283--17290},
  year={2022},
  publisher={IEEE}
}

@article{lu2024videomambapro,
  title={Videomambapro: A leap forward for mamba in video understanding},
  author={Lu, Hui and Salah, Albert Ali and Poppe, Ronald},
  journal={arXiv preprint arXiv:2406.19006},
  year={2024}
}

@inproceedings{chen2023run,
  title={Run, don't walk: chasing higher FLOPS for faster neural networks},
  author={Chen, Jierun and Kao, Shiu-hong and He, Hao and Zhuo, Weipeng and Wen, Song and Lee, Chul-Ho and Chan, S-H Gary},
  booktitle={Proceedings of the IEEE/CVF Conference on Computer Vision and Pattern Recognition},
  pages={12021--12031},
  year={2023}
}

@article{zhao2024cobra,
  title={Cobra: Extending mamba to multi-modal large language model for efficient inference},
  author={Zhao, Han and Zhang, Min and Zhao, Wei and Ding, Pengxiang and Huang, Siteng and Wang, Donglin},
  journal={arXiv preprint arXiv:2403.14520},
  year={2024}
}

@article{allan2020endovis18,
  title={2018 robotic scene segmentation challenge},
  author={Allan, Max and Kondo, Satoshi and Bodenstedt, Sebastian and Leger, Stefan and Kadkhodamohammadi, Rahim and Luengo, Imanol and Fuentes, Felix and Flouty, Evangello and Mohammed, Ahmed and Pedersen, Marius and others},
  journal={arXiv preprint arXiv:2001.11190},
  year={2020}
}

@article{he2024panmamba,
  title={Pan-mamba: Effective pan-sharpening with state space model},
  author={He, Xuanhua and Cao, Ke and Yan, Keyu and Li, Rui and Xie, Chengjun and Zhang, Jie and Zhou, Man},
  journal={arXiv preprint arXiv:2402.12192},
  year={2024}
}

@article{qiao2024vlmamba,
  title={Vl-mamba: Exploring state space models for multimodal learning},
  author={Qiao, Yanyuan and Yu, Zheng and Guo, Longteng and Chen, Sihan and Zhao, Zijia and Sun, Mingzhen and Wu, Qi and Liu, Jing},
  journal={arXiv preprint arXiv:2403.13600},
  year={2024}
}

@article{li2024spikemba,
  title={Spikemba: Multi-modal spiking saliency mamba for temporal video grounding},
  author={Li, Wenrui and Hong, Xiaopeng and Fan, Xiaopeng},
  journal={arXiv preprint arXiv:2404.01174},
  year={2024}
}

@misc{liu2024vmambavisualstatespace,
      title={VMamba: Visual State Space Model}, 
      author={Yue Liu and Yunjie Tian and Yuzhong Zhao and Hongtian Yu and Lingxi Xie and Yaowei Wang and Qixiang Ye and Yunfan Liu},
      year={2024},
      eprint={2401.10166},
      archivePrefix={arXiv},
      primaryClass={cs.CV},
      url={https://arxiv.org/abs/2401.10166}, 
}

@inproceedings{ma2021ldpolypvideo,
  title={LDPolypVideo benchmark: a large-scale colonoscopy video dataset of diverse polyps},
  author={Ma, Yiting and Chen, Xuejin and Cheng, Kai and Li, Yang and Sun, Bin},
  booktitle={International Conference on Medical Image Computing and Computer-Assisted Intervention},
  pages={387--396},
  year={2021},
  organization={Springer}
}

@inproceedings{resnet,
	year = 2016,
	author = {Kaiming He and Xiangyu Zhang and Shaoqing Ren and Jian Sun},
	title = {Deep Residual Learning for Image Recognition},
	booktitle = {Proceedings of the IEEE Conference on Computer Vision and Pattern Recognition},
	pages = {770-778}
}

@article{liu2022tcnet,
  title={Temporal consistency learning of inter-frames for video super-resolution},
  author={Liu, Meiqin and Jin, Shuo and Yao, Chao and Lin, Chunyu and Zhao, Yao},
  journal={IEEE Transactions on Circuits and Systems for Video Technology},
  volume={33},
  number={4},
  pages={1507--1520},
  year={2022},
  publisher={IEEE}
}

@inproceedings{chan2021basicvsr,
  title={Basicvsr: The search for essential components in video super-resolution and beyond},
  author={Chan, Kelvin CK and Wang, Xintao and Yu, Ke and Dong, Chao and Loy, Chen Change},
  booktitle={Proceedings of the IEEE/CVF Conference on Computer Vision and Pattern Recognition},
  pages={4947--4956},
  year={2021}
}

@article{shi2022rethinking,
  title={Rethinking alignment in video super-resolution transformers},
  author={Shi, Shuwei and Gu, Jinjin and Xie, Liangbin and Wang, Xintao and Yang, Yujiu and Dong, Chao},
  journal={Advances in Neural Information Processing Systems},
  volume={35},
  pages={36081--36093},
  year={2022}
}

@inproceedings{chan2022basicvsr++,
  title={Basicvsr++: Improving video super-resolution with enhanced propagation and alignment},
  author={Chan, Kelvin CK and Zhou, Shangchen and Xu, Xiangyu and Loy, Chen Change},
  booktitle={Proceedings of the IEEE/CVF Conference on Computer Vision and Pattern Recognition},
  pages={5972--5981},
  year={2022}
}

@article{cao2021vsrt,
  title={Video super-resolution transformer},
  author={Cao, Jiezhang and Li, Yawei and Zhang, Kai and Van Gool, Luc},
  journal={arXiv preprint arXiv:2106.06847},
  year={2021}
}

@inproceedings{wang2019edvr,
  title={Edvr: Video restoration with enhanced deformable convolutional networks},
  author={Wang, Xintao and Chan, Kelvin CK and Yu, Ke and Dong, Chao and Change Loy, Chen},
  booktitle={Proceedings of the IEEE/CVF Conference on Computer Vision and Pattern Recognition Workshops},
  pages={0--0},
  year={2019}
}

@article{liang2022rvrt,
  title={Recurrent video restoration transformer with guided deformable attention},
  author={Liang, Jingyun and Fan, Yuchen and Xiang, Xiaoyu and Ranjan, Rakesh and Ilg, Eddy and Green, Simon and Cao, Jiezhang and Zhang, Kai and Timofte, Radu and Gool, Luc V},
  journal={Advances in Neural Information Processing Systems},
  volume={35},
  pages={378--393},
  year={2022}
}

@inproceedings{yi2019progressive,
  title={Progressive fusion video super-resolution network via exploiting non-local spatio-temporal correlations},
  author={Yi, Peng and Wang, Zhongyuan and Jiang, Kui and Jiang, Junjun and Ma, Jiayi},
  booktitle={Proceedings of the IEEE/CVF International Conference on Computer Vision},
  year={2019}
}

@article{qiu2023rethinkingsrmedical,
  title={Rethinking Dual-Stream Super-Resolution Semantic Learning in Medical Image Segmentation},
  author={Qiu, Zhongxi and Hu, Yan and Chen, Xiaoshan and Zeng, Dan and Hu, Qingyong and Liu, Jiang},
  journal={IEEE Transactions on Pattern Analysis and Machine Intelligence},
  year={2023},
  publisher={IEEE}
}

@inproceedings{krishnan1998intestinal,
  title={Intestinal abnormality detection from endoscopic images},
  author={Krishnan, SM and Yang, X and Chan, KL and Kumar, S and Goh, PMY},
  booktitle={Proceedings of the 20th Annual International Conference of the IEEE Engineering in Medicine and Biology Society. Vol. 20 Biomedical Engineering Towards the Year 2000 and Beyond (Cat. No. 98CH36286)},
  volume={2},
  pages={895--898},
  year={1998},
  organization={IEEE}
}

@article{yuan2024remamba,
  title={ReMamba: Equip Mamba with Effective Long-Sequence Modeling},
  author={Yuan, Danlong and Liu, Jiahao and Li, Bei and Zhang, Huishuai and Wang, Jingang and Cai, Xunliang and Zhao, Dongyan},
  journal={arXiv preprint arXiv:2408.15496},
  year={2024}
}

@inproceedings{guo2024mambair,
  title={Mambair: A simple baseline for image restoration with state-space model},
  author={Guo, Hang and Li, Jinmin and Dai, Tao and Ouyang, Zhihao and Ren, Xudong and Xia, Shu-Tao},
  booktitle={European Conference on Computer Vision},
  pages={222--241},
  year={2025},
  organization={Springer}
}

@article{chow2016review,
  title={Review of medical image quality assessment},
  author={Chow, Li Sze and Paramesran, Raveendran},
  journal={Biomedical Signal Processing and Control},
  volume={27},
  pages={145--154},
  year={2016},
  publisher={Elsevier}
}

@article{peters2001image,
  title={Image-guided surgery: from X-rays to virtual reality},
  author={Peters, Terry M},
  journal={Computer Methods in Biomechanics and Biomedical Engineering},
  volume={4},
  number={1},
  pages={27--57},
  year={2001},
  publisher={Taylor \& Francis}
}

@inproceedings{chollet2017xception,
  title={Xception: Deep learning with depthwise separable convolutions},
  author={Chollet, Fran{\c{c}}ois},
  booktitle={Proceedings of the IEEE Conference on Computer Vision and Pattern Recognition},
  pages={1251--1258},
  year={2017}
}

@inproceedings{cataract,
  author    = {Klaus Schoeffmann and
               Mario Taschwer and
               Stephanie Sarny and
               Bernd M{\"{u}}nzer and
               Manfred J{\"{u}}rgen Primus and
               Doris Putzgruber},
  editor    = {Pablo C{\'{e}}sar and
               Michael Zink and
               Niall Murray},
  title     = {Cataract-101: video dataset of 101 cataract surgeries},
  booktitle = {Proceedings of the 9th {ACM} Multimedia Systems Conference},
  pages     = {421--425},
  publisher = {{ACM}},
  year      = {2018},
  url       = {https://doi.org/10.1145/3204949.3208137},
  doi       = {10.1145/3204949.3208137},
  bibsource = {dblp computer science bibliography, https://dblp.org}
}

@article{lucas2019generative,
  title={Generative adversarial networks and perceptual losses for video super-resolution},
  author={Lucas, Alice and Lopez-Tapia, Santiago and Molina, Rafael and Katsaggelos, Aggelos K},
  journal={IEEE Transactions on Image Processing},
  volume={28},
  number={7},
  pages={3312--3327},
  year={2019},
  publisher={IEEE}
}

@inproceedings{yan2019frame,
  title={Frame and feature-context video super-resolution},
  author={Yan, Bo and Lin, Chuming and Tan, Weimin},
  booktitle={Annual {AAAI} Conference on Artificial Intelligence},
  year={2019}
}

@inproceedings{liu2021swin,
  title={Swin transformer: Hierarchical vision transformer using shifted windows},
  author={Liu, Ze and Lin, Yutong and Cao, Yue and Hu, Han and Wei, Yixuan and Zhang, Zheng and Lin, Stephen and Guo, Baining},
  booktitle={Proceedings of the IEEE/CVF International Conference on Computer Vision},
  year={2021}
}

@inproceedings{ranjan2017optical,
  title={Optical flow estimation using a spatial pyramid network},
  author={Ranjan, Anurag and Black, Michael J},
  booktitle={Proceedings of the IEEE/CVF Conference on Computer Vision and Pattern Recognition},
  year={2017}
}

@inproceedings{nah2019ntire,
  title={Ntire 2019 challenge on video deblurring and super-resolution: Dataset and study},
  author={Nah, Seungjun and Baik, Sungyong and Hong, Seokil and Moon, Gyeongsik and Son, Sanghyun and Timofte, Radu and Mu Lee, Kyoung},
  booktitle={Proceedings of the IEEE/CVF Conference on Computer Vision and Pattern Recognition Workshops},
  year={2019}
}

@article{gu2023mamba,
  title={Mamba: Linear-time sequence modeling with selective state spaces},
  author={Gu, Albert and Dao, Tri},
  journal={arXiv preprint arXiv:2312.00752},
  year={2023}
}

@article{dao2024mamba2,
  title={Transformers are SSMs: Generalized models and efficient algorithms through structured state space duality},
  author={Dao, Tri and Gu, Albert},
  journal={arXiv preprint arXiv:2405.21060},
  year={2024}
}

@inproceedings{dai2017deformable,
  title={Deformable convolutional networks},
  author={Dai, Jifeng and Qi, Haozhi and Xiong, Yuwen and Li, Yi and Zhang, Guodong and Hu, Han and Wei, Yichen},
  booktitle={Proceedings of the IEEE/CVF International Conference on Computer Vision},
  year={2017}
}

@article{wang2019deformable,
  title={Deformable non-local network for video super-resolution},
  author={Wang, Hua and Su, Dewei and Liu, Chuangchuang and Jin, Longcun and Sun, Xianfang and Peng, Xinyi},
  journal={IEEE Access},
  volume={7},
  pages={177734--177744},
  year={2019},
  publisher={IEEE}
}

@article{xue2019video,
  title={Video enhancement with task-oriented flow},
  author={Xue, Tianfan and Chen, Baian and Wu, Jiajun and Wei, Donglai and Freeman, William T},
  journal={International Journal of Computer Vision},
  volume={127},
  number={8},
  pages={1106--1125},
  year={2019},
  publisher={Springer}
}

@inproceedings{kim2018spatio,
  title={Spatio-temporal transformer network for video restoration},
  author={Kim, Tae Hyun and Sajjadi, Mehdi SM and Hirsch, Michael and Scholkopf, Bernhard},
  booktitle={European Conference on Computer Vision},
  year={2018}
}

@inproceedings{sajjadi2018frame,
  title={Frame-recurrent video super-resolution},
  author={Sajjadi, Mehdi SM and Vemulapalli, Raviteja and Brown, Matthew},
  booktitle={Proceedings of the IEEE/CVF Conference on Computer Vision and Pattern Recognition},
  year={2018}
}

@inproceedings{charbonnier1994two,
  title={Two deterministic half-quadratic regularization algorithms for computed imaging},
  author={Charbonnier, Pierre and Blanc-Feraud, Laure and Aubert, Gilles and Barlaud, Michel},
  booktitle={Proceedings of the International Conference on Image Processing},
  volume={2},
  pages={168--172},
  year={1994},
  organization={IEEE}
}

@article{Dosovitskiy2021ViT,
  title={An Image is Worth 16x16 Words: Transformers for Image Recognition at Scale},
  author={Dosovitskiy, Alexey and Beyer, Lucas and Kolesnikov, Alexander and Weissenborn, Dirk and Zhai, Xiaohua and Unterthiner, Thomas and  Dehghani, Mostafa and Minderer, Matthias and Heigold, Georg and Gelly, Sylvain and Uszkoreit, Jakob and Houlsby, Neil},
  booktitle={ICLR},
  year={2021}
}

@article{gu2021efficiently,
  title={Efficiently modeling long sequences with structured state spaces},
  author={Gu, Albert and Goel, Karan and R{\'e}, Christopher},
  journal={arXiv preprint arXiv:2111.00396},
  year={2021}
}

@article{nguyen2022s4nd,
  title={S4nd: Modeling images and videos as multidimensional signals with state spaces},
  author={Nguyen, Eric and Goel, Karan and Gu, Albert and Downs, Gordon and Shah, Preey and Dao, Tri and Baccus, Stephen and R{\'e}, Christopher},
  journal={Advances in Neural Information Processing Systems},
  volume={35},
  pages={2846--2861},
  year={2022}
}

@inproceedings{ding2022scaling,
  title={Scaling up your kernels to 31x31: Revisiting large kernel design in cnns},
  author={Ding, Xiaohan and Zhang, Xiangyu and Han, Jungong and Ding, Guiguang},
  booktitle={Proceedings of the IEEE/CVF Conference on Computer Vision and Pattern Recognition},
  pages={11963--11975},
  year={2022}
}

@inproceedings{liu2022convnet,
  title={A convnet for the 2020s},
  author={Liu, Zhuang and Mao, Hanzi and Wu, Chao-Yuan and Feichtenhofer, Christoph and Darrell, Trevor and Xie, Saining},
  booktitle={Proceedings of the IEEE/CVF Conference on Computer Vision and Pattern Recognition},
  pages={11976--11986},
  year={2022}
}

@inproceedings{wang2023internimage,
  title={Internimage: Exploring large-scale vision foundation models with deformable convolutions},
  author={Wang, Wenhai and Dai, Jifeng and Chen, Zhe and Huang, Zhenhang and Li, Zhiqi and Zhu, Xizhou and Hu, Xiaowei and Lu, Tong and Lu, Lewei and Li, Hongsheng and others},
  booktitle={Proceedings of the IEEE/CVF Conference on Computer Vision and Pattern Recognition},
  pages={14408--14419},
  year={2023}
}

@inproceedings{xu2024iart,
  title={Enhancing Video Super-Resolution via Implicit Resampling-based Alignment},
  author={Xu, Kai and Yu, Ziwei and Wang, Xin and Mi, Michael Bi and Yao, Angela},
  booktitle={Proceedings of the IEEE/CVF Conference on Computer Vision and Pattern Recognition},
  pages={2546--2555},
  year={2024}
}
